%% file: main.tex
\documentclass{article}
\usepackage{spconf,amsmath,graphicx}

\usepackage{enumitem}
\setlist{nosep, leftmargin=14pt}
\usepackage{pgfplots}
\pgfplotsset{compat=1.16}
\usepackage{pgfplotstable}
\usepackage{hyperref}
\usepackage{amsmath}
\usepackage{algorithm}
\usepackage{algorithmic}
\usepgfplotslibrary{patchplots}
\usepackage{pgffor}
\usepackage{tikz}
\usepackage{amssymb}

\newcommand{\ie}[1]{\emph{i.e.}}
\newcommand{\eg}[1]{\emph{e.g.}}
\DeclareMathOperator*{\argmax}{arg\,max}

\DeclareMathOperator*{\IOU}{IOU}
\newcommand{\rv}[1]{\mathbf{#1}}

\title{Bounding Box Priors for Cell Detection with Point Annotations}

\name{Hari Om Aggrawal$^{1}$ \qquad Dipam Goswami$^{2}$ \qquad Vinti Agarwal$^{2}$}
\address{$^{1}$ Independent Researcher, India \sthanks{Previously affiliated with Institute of Mathematics and Image Computing, University of Luebeck, Germany. Email: hariom85@gmail.com}\\
     $^{2}$ Birla Institute of Technology and Science, Pilani, India}
\begin{document}
%
\maketitle
\begin{abstract}
The size of an individual cell type, such as a red blood cell, does not vary much among humans.
We use this knowledge as a prior for classifying and detecting cells in images with only a few ground truth bounding box annotations, while most of the cells are annotated with points.
This setting leads to weakly semi-supervised learning. We propose replacing points with either stochastic (ST) boxes or bounding box predictions during the training process. The proposed  ``mean-IOU" ST box maximizes the overlap with all the boxes belonging to the sample space with a class-specific approximated prior probability distribution of bounding boxes. Our method trains with both box- and point-labelled images in conjunction, unlike the existing methods, which train first with box- and then point-labelled images.
In the most challenging setting, when only $5\%$ images are box-labelled, quantitative experiments on a urine dataset show that our one-stage method outperforms two-stage methods by $5.56$ mAP. Furthermore, we suggest an approach that partially answers ``how many box-labelled annotations are necessary?" before training a machine learning model.

\end{abstract}

\input{introduction}

\input{method}

\input{results}

\input{conclusion}


\bibliographystyle{IEEEbib}
\bibliography{refs}

\end{document}

%% file: introduction.tex
\section{Introduction}
\label{sec:intro}

Microscopic urinalysis is a gold-standard diagnostic test to detect urinary tract infections and kidney disorders \cite{liang2018object}. Anemia is usually determined by the microscopic examination of the blood \cite{walker1990clinical}. Urine and blood contain various cells that must be accurately detected and classified for medical diagnosis. 

Deep learning-based supervised models are state-of-the-art for detecting objects. But, they are hungry for high-quality labelled training data. Numerous urine and blood samples can be easily collected in labs, but their annotation is time-consuming and requires extensive medical expert knowledge. Therefore, weakly and/or semi-supervised models are preferable if they perform similarly to a supervised model.

The shape, size, and features of a particular cell do not vary much among humans. However, their appearances differ a lot in microscopic digital images. Improper focus adjustments, lighting conditions, and the quality of image sensors and lenses affect the true features of a cell. Moreover, images contain just two-dimensional (2D) projections of a 3D cell. Therefore, many cell instances are required to train a deep-learning model for classification.

On the contrary, the size of bounding boxes covering cells does not change much, as we notice by drawing joint probability density functions (pdf) of the width and height of bounding boxes; see Fig.~\ref{fig:pdf}. In urinary-sediment dataset (USD) \cite{liang2018object}, the erythrocyte and leukocyte classes are mainly square and have a very narrow distribution. Other classes, except epithelial nuclei, are rectangular and have wider distributions due to both the size and the orientation of cells. 

Due to the fewer deviations in size, we hypothesize that the bounding box regression requires fewer ground truth data samples than the samples required for classification. To validate our hypothesis, we trained RetinaNet \cite{lin2017focal} with only 5\% box-labelled images of the USD. We observed that the model predicts 73\% of objects accurately at a confidence score 0.05, even with only a very few bounding box annotations.

This paper proposes a weakly semi-supervised method that trains with a few box-labelled annotations. Most cells are annotated with points following the annotation strategy in \cite{papadopoulos2017training}. The point is placed close to the center of the object. 

The existing weakly semi-supervised approaches use two-stage training. A teacher model trains with only box-labelled images to generate pseudo boxes for the point-labelled instances. Afterward, a student model trains with both box- and pseudo-box-labelled images \cite{chen2021points, zhang2022group}. The performance of the teacher model highly depends on the variations in and quality of box-labelled annotations. Moreover, for two-stage methods, it is challenging to handle datasets having both bounding boxes and point annotations in the same image.

We propose a one-stage training method that jointly trains with both box- and point-labelled images. Points are replaced with a stochastic (ST) box or a prediction from the box regression subnet. The proposed ``mean-IOU'' ST box utilizes the class-specific joint pdf approximated with box-labelled instances available in the dataset.

Sec.~\ref{sec:method} describes a point-to-box selection algorithm and a novel loss function. In Sec.~\ref{sec:results}, we compare one-stage and two-stage methods on urine and blood datasets.

%% file: method.tex
\input{plot_jointpdf}

\section{Method}\label{sec:method}

In this section, we present a one-stage, weakly semi-supervised learning strategy that utilizes both strong and weak annotations jointly in the form of ground-truth bounding boxes and points, respectively, during training.

We describe our method by considering RetinaNet~\cite{lin2017focal} as a base model. However, the proposed method is applicable to any anchor-based detection models with mild modifications. 

RetinaNet consists of a backbone network and two subnetworks; classification and box regression. The backbone computes feature maps. The pre-defined object proposals, called anchors, are defined at each spatial position of the feature maps. A set of anchors are assigned to each ground-truth object box, called foreground anchors using an IOU threshold of $0.5$ and to the background if their IOU is in $[0, 0.4)$. The classification subnet predicts the probability of object presence at each spatial position for each of the anchors and object classes. The box regression subnet regresses the offset from each anchor box to a nearby ground-truth (GT) object box.

The anchor assignment, as described above, is one of the crucial steps; wrong assignments lead to lower precision. For point annotations, object boundaries are not available; hence anchor assignment is a complex task. A naive approach could assign an anchor as foreground if it contains the annotated point. However, this would select many anchors that cover spatial regions far beyond the ground-truth object and produce many false positives.

This paper proposes a method to build a bounding box around each point annotation during training. We refer to them as stochastic (ST) boxes because they are defined from a class-specific probability distribution of bounding boxes. These boxes change based on the box regression subnet's output at every iteration. Our idea is simple to implement and requires minor changes to the base RetinaNet model. 

\subsection{Point to Stochastic Box}

During training, one can utilize the bounding box predictions from the box regression subnet for objects having point annotations. However, there are two major issues that need attention. First, how to handle inaccurate predictions at the initial stage of training?; second, how to ensure at least one credible prediction corresponding to each point annotation to avoid considering the object as a background? 

We propose algorithm \ref{algo:stbox} that uses predictions only if they are reliable; otherwise, build a ST box utilizing a class-specific joint pdf $f_c$ of width $\rv{w}$ and height $\rv{h}$ where $c$ denotes a class. We measure $f_c(w,h)$ from the box-labelled annotations present in the train set using a kernel density estimator with Gaussian kernel; see Fig.~\ref{fig:pdf} for joint pdf and the estimates of ST boxes. We discuss two choices for ST boxes: \\
\textbf{S1) Mean:} We define a ``mean" ST box of width $\hat{w}$ and height $\hat{h}$ with their expected values, \ie, 
\[
    \hat{w} = \int w f_c(w) dw, \quad \hat{h} = \int h f_c(h) dh
\]
where $f_c(w)$ and $f_c(h)$ are the marginal pdf of width and height, respectively. The center of the ST box is defined at the annotated point for an object. \\
\textbf{S2) Mean-IOU:} The ``mean" ST box does not capture the cross-correlation between the width and height, whereas the proposed ``mean-IOU" ST box does. It is defined as a box that has maximum overlap with all the boxes belonging to the sample space of bounding boxes, which is a solution to the optimization problem
\begin{equation}\label{eq:mean_iou}
    (\tilde{w}, \tilde{h}) = \argmax_{x, y} \int \IOU\Big(B(x,y), B(w,h)\Big) \ f_c(w,h) \ dw \ dh
\end{equation}
where $B(w,h)$ denotes a box of width $w$ and height $h$ whose center is at origin without loss of a generality and $\IOU(A,B)$ denotes the IOU between box A and B.

From the law of the unconscious statistician (LOTUS), the objective function in \eqref{eq:mean_iou} measures the expected value of IOU. Therefore, we name the measure ``mean-IOU".

We solve \eqref{eq:mean_iou} by discretizing the integral on $100 \times 100$ rectangular grid and optimizing with the gradient-free patternsearch algorithm. The initial estimate is set to the ``mean" ST box for faster convergence. We further reduce the solution space by setting the lower and upper bound with $\mu \pm 5 \sigma$ where $\mu$ is the mean, and $\sigma$ is the standard deviation of width and height variables. We verified that the final estimate converges to the same optima irrespective of any initial estimate. 

Note, \eqref{eq:mean_iou} needs to be solved only once at the start of the training process and it is quick to solve; it took $0.04$sec. (on average) per class for the USD dataset on MATLAB running on the Apple M1 Pro (2021) without GPU support.

\subsubsection{Algorithm:}

We assume that any credible prediction for an object will overlap with the 
``mean-IOU" ST box. Hence, a prediction is assigned to a point if IOU with the ``mean-IOU" ST box is above a threshold $\tau_{iou}$; otherwise ``mean-IOU" ST box is assigned to the point. In this way, we avoid a point-annotated object being treated as a background during training.

Usually, more than one prediction overlaps with the ``mean-IOU" ST box. Hence, we need an extra filtering step that assigns one ``best" prediction to a point. Motivated from \cite{papadopoulos2017training}, the ``best" prediction maximizes the score
\[
s(B) = P(B) \exp{(-\|p - u\|_2^2)}
\]
where $P(B)$ denotes the probability of object presence, \ie, the confidence score from the classification subnet, $u = (u_x, u_y)$ is the center of the box $B$, and $p = (p_x, p_y)$ is the location of point annotation. $\|p - u\|_2^2$ measures the distance from the center of prediction to the annotated point.

The above-described scheme might provide unreliable boxes at the initial stage of training. Therefore, we select only high confidence score predictions whose score is $\geq \tau_{s}$. In our experiments, we use $\tau_{s} = 0.2$ and  $\tau_{iou} = 0.5$.

\renewcommand{\algorithmicrequire}{\textbf{Input:}}
\renewcommand{\algorithmicensure}{\textbf{Output:}}
\begin{algorithm}[t]
	\caption{Point to box selection strategy.}%
	\label{algo:stbox}
	\begin{algorithmic}[1]
		\REQUIRE{Point annotation $p = (p_x, p_y)$, mean-IOU ST box $S$ centered at $p$, predicted boxes from box regression subnet $D = \{d_1, \dots, d_N\}$, $\tau_{iou} > 0$, $\tau_{s} > 0$.}
		\ENSURE{a box to replace the point for assigning anchors}
		\STATE{$D_t = \{ d_i | d_i \in D,  P(d_i) \geq \tau_{s}, \IOU(S,d_i) \geq \tau_{iou}\}$}
		\IF{$D_t = \emptyset$}
		\RETURN{S}
		\ELSE
		\STATE{$f(d_i) = P(d_i) \exp(-\|p - u_i\|_2^2) \ \forall d_i \in D_t$, \ $u_i$ is the center of $d_i^{th}$ box}
		\STATE{$d_{best} = d_n \ \text{where} \ n = \argmax_{k} \{f(d_k)\}$}
		\RETURN{$d_{best}$}
		\ENDIF
	\end{algorithmic}
\end{algorithm}

\subsection{Loss Functions}
Following the RetinaNet anchor assignment strategy for GT boxes, a set of anchors are assigned to selected boxes from Algo~\ref{algo:stbox} with probability one to minimize the cross-entropy loss function to learn the classification subnet's parameters.

The box regression subnet regresses the offset from each foreground anchor to only nearby GT boxes, not the selected boxes. These boxes are just an approximation; the regression with respect to them performed poorly in our experiments. The annotated points are the ground truth; therefore, the distance from the anchor's center to the point is minimized.

Point annotations are noisier than bounding boxes \cite{papadopoulos2017training}; annotating a point at the exact center of the object is a cumbersome task. Therefore, the box regression loss function, denoted by $L$, downweight the contribution from point annotations by a factor of $(1 + \beta)$, $\beta \geq 0$;
\begin{equation}
L = \frac{1}{N_g} \sum_{i=1}^{N_g} \|\hat{t}_i - (t_g)_i\|_1 + \frac{1}{N_p} \sum_{j=1}^{N_p} \frac{\|\hat{t}_j - (t_p)_j\|_1}{1 + \beta}
\end{equation}
where $\hat{t}$ are the box subnet predictions, $(t_g)_i \in \mathbb{R}^4$ and $(t_p)_i \in \mathbb{R}^2$ are the offset between a foreground anchor and the associated GT box or the point annotation, respectively. Note that, $t_p$ are just the centers' offset. $N$ denotes the number of foreground anchors and $\|.\|_1$,  the L1 norm.

The hyperparameter $\beta$ can be considered a measure of uncertainty in the points; the lesser the uncertainty, the smaller the $\beta$. The noise analysis in \cite{papadopoulos2017training} says that the error distance between the true center and the point annotation increases proportionally to the object area. Therefore, we set $\beta = \alpha/\sqrt{A_c}, \alpha \geq 0,$ where $A_c$ is the mean area of bounding boxes associated with the $c^{th}$ class.

%% file: plot_jointpdf.tex
\pgfplotsset{
colormap/viridis,
patch type=bilinear
}

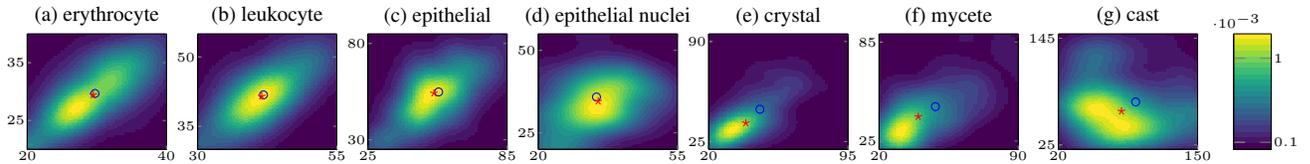
\begin{figure*}[t!]
\centering
\def\first{cast}
\foreach \c/\n/\m/\xmin/\xmax in {eryth/erythrocyte/a/20/40,leuko/leukocyte/b/30/55,epith/epithelial/c/25/85,epithn/epithelial nuclei/d/20/55,cryst/crystal/e/20/95,mycete/mycete/f/20/90,cast/cast/g/20/150}{
   \begin{minipage}[b]{0.14\columnwidth}%
   	       \pgfplotstableread[col sep=comma]{data/USD_\c_joint_pdf.csv}\loadedtable
            \begin{tikzpicture}[font=\footnotesize, inner sep=0pt, outer sep=1pt]
              \begin{axis}[
              	view={0}{90},
              	title={(\m) \n},
              	title style={yshift=-4.0pt,},
              	scale=0.27,
              	mark size=1.5pt,
              	xmin=\xmin,xmax=\xmax,
              	ymin=\xmin,ymax=\xmax,
              	xtick={\xmin,\xmax},
                ytick={\xmin+5,\xmax-5},
              	tick label style={font=\tiny},
              	max space between ticks=100,
              	try min ticks=2,
              	tickwidth=0.05cm,
              	table/col sep=comma,
              	\ifx\first\n colorbar \fi,
              	colorbar style={
              		ytick={0.0001, 0.001},
              	}
              ]
              \addplot3[contour filled={number = 20}] table[x=x1, y=x2, z=\c]{\loadedtable};
              \addplot+[only marks, blue, mark=o] table[x=\c_x,y=\c_y]{data/USD_mu.csv};
              \addplot+[only marks, red, mark=star] table[x=\c_x,y=\c_y]{data/USD_miou.csv};
              \end{axis}
            \end{tikzpicture}
   \end{minipage}\hfil%
}
\caption{Approximated joint pdf of width and height of bounding boxes of all seven classes in USD-50 with only $20\%$ box-labelled instances and the estimates of ``mean'' (blue circle) and ``mean-IOU'' (red star) ST boxes. For highly skewed distributions, \eg~for crystal, mycete, and cast, training with ``mean-IOU'' tends to produce less false positives than with the ``mean''.}
\label{fig:pdf}
\end{figure*}

%% file: results.tex
\section{Experiments and Results}
\label{sec:results}

\input{plot_results}

We present experimental results on two datasets; USD and BCCD \cite{bccd}. We randomly select $50\%$ of images from USD train set and form a new set USD-50 for training; the trained models with USD and USD-50 attain almost the same mean average precision (mAP) on the val set. USD-50 has 16,294 cell instances from 7 classes. BCCD training set only has 2805 cell instances from 3 classes. The ground truth class label and bounding box are available for each cell instance.

We partition the train set into two mutually exclusive subsets; well-labelled and weakly-labelled sets. In the weakly-labelled set, we replace a bounding box with a point.\\
\textbf{Points generation:} Following the point-annotations error analysis in \cite{papadopoulos2017training}, we introduce noise to the true center of the object bounding boxes. In USD, $99.9\%$ object's area, $A \in [18^2, 198^2]$ pixels for which the mean error-distances $\mu$ vary from 5 to 17 pixels as the object's area increases; see Fig.4 in \cite{papadopoulos2017training}. To mimic this behaviour, we assume that the noise follows a normal distribution whose mean $\mu$ is a linear function of $\sqrt{A}$ and the standard deviation is $3$ pixels. This setting is highly noisy for small-size cells, \eg, the annotated center of erythrocyte cells in USD is shifted on average by $40\%$ with respect to the cell's average width and height.

For training, we optimize parameters with SGD for 50 epochs with an initial learning rate 0.01, weight decay factor of $10^{-4}$, and momentum 0.9, on one Tesla V100-PCIE-32GB GPU. The learning rate is reduced when the loss has stopped improving. We obtain the best training stability and generalization performance with a batch size of 8 for USD and 2 for BCCD. All experiments are run at a fixed random seed. The ResNet-50 backbone is pre-trained on Imagenet. Both USD and BCCD have a high-class imbalance; therefore, we use the class-balance loss strategy from \cite{cui2019class} with $\beta= 0.99$.

We compare the proposed one-stage method with a baseline and a so-called two-stage method upper bound. The baseline model trains only with the well-labelled training set. We use the baseline model to generate pseudo boxes for the weakly-labelled set and then train with both well-labelled and pseudo box-labelled images for a two-stage method. We assume that a best-performing first stage of a two-stage method will predict the pseudo boxes whose IOU with the ground truth bounding boxes of weakly-labelled set is higher than $0.5$. This assumption is followed to generate pseudo boxes; the second-stage performance is reported as an upper bound for the two-stage method.

The results are summarized in Fig.~\ref{fig:results}. Two-stage method upper bounds can only be constructed for point annotations without noise. Therefore, we also report results for the one-stage method without noise for a fair comparison between methods. We report results only with mean-IOU ST box.

In the most challenging setting, when only $5\%$ images are well-labelled, and the rest $95\%$ images are weakly-labelled, the one-stage method outperforms the two-stage method by 17.52 mAP without noise and 5.56 mAP with noise for USD; similar trend also follows for BCCD. As expected, the performance gap decreases with an increase in the number of box-labelled instances in the training set.

We observed the correlation between the baseline mAP and the KL divergence of pdfs; see Fig.~\ref{fig:results} for details. Noting that, we advise annotating bounding boxes until the pdf of bounding boxes starts to converge, annotating the rest of the objects with points, and using a weakly semi-supervised method for training.

%% file: plot_results.tex
\pgfplotsset{
compat=1.7,
	y axis style/.style={
	yticklabel style=#1,
	ylabel style=#1,
	y axis line style=#1,
	ytick style=#1
}
}

\begin{figure*}[ht!]
    \centering
    \begin{minipage}[b]{0.60\linewidth}
    \pgfplotstableread[col sep=comma]{data/USD.csv}\loadedtable
    \begin{tikzpicture}[font=\footnotesize,inner sep=2pt, outer sep=0pt]
      \begin{axis}[
        axis y line=left,
        scale = 0.7,
        title = {(a) USD-50},
        title style={yshift=-3.0pt,},
        table/col sep=comma,
        axis lines = left,
        xmin=0, xmax=55, ymin=0, ymax=90,
        xtick={5, 10, 20, 30, 40, 50},
        xlabel = $\%$ of well-labelled (wl) images, ylabel = average mAP,
        minor tick num=0,
        grid=both,
        grid style={line width=.1pt, draw=gray!10},
        mark size=1.5pt,
        legend entries = {baseline, two-stage upper bound \\ (without noise), one-stage (ours) \\ (without noise \, $\alpha = 0$), one-stage (ours) \\ (with noise \, $\alpha = 10$), mAP at $100\%$ \\ well-labelled images },
        legend style={draw=none,inner sep=0pt, outer sep=0pt,at={(1.65,1)},
anchor=north,cells={align=left}},
        legend cell align={left},
      ]
      \addplot[blue,mark=o] table[x=wl, y=B]{\loadedtable};
      \addplot[gray, mark=asterisk] table[x=wl, y=P]{\loadedtable};
      \addplot[green, mark=otimes] table[x=wl, y=S]{\loadedtable};
      \addplot[orange, mark=*] table[x=wl, y=SN]{\loadedtable};
      \addplot[mark=none, black, dashed] coordinates {(0,79.56) (52,79.56)};
      \end{axis}
      \begin{axis}[
	      xmin=0, xmax=55,scale = 0.7,
	      ymax=1,
	      hide x axis,
	      axis y line=right,
	      ylabel={KL div.},
	      ylabel style={yshift=7.5pt,},
	      table/col sep=comma,
	      y axis style=red,
	      legend entries = {KL divergence},
	      legend style={draw=none,inner sep=0pt, outer sep=0pt,at={(1.55,0.15)},
	      	anchor=north,cells={align=left}},
	      legend cell align={left},
      ]
      \addplot[red, dashed, mark=otimes, mark size=1pt] table[x=x, y=y]{data/USD_KLdiv.csv};
      \end{axis}
    \end{tikzpicture}
    \end{minipage}
    \begin{minipage}[b]{0.39\linewidth}
    \pgfplotstableread[col sep=comma]{data/BCCD.csv}\loadedtable
    \begin{tikzpicture}[font=\footnotesize,inner sep=2pt, outer sep=0pt]
      \begin{axis}[
        scale = 0.7,
        title = {(b) BCCD},
        title style={yshift=-3.0pt,},
        table/col sep=comma,
        axis lines = left,
        mark size=1.5pt,
        xmin=0, xmax=55, ymin=0, ymax=90,
        xtick={5, 10, 20, 30, 40, 50},
        xlabel = $\%$ of well-labelled (wl) images, ylabel = average mAP,
        minor tick num=0,
        grid=both,
        grid style={line width=.1pt, draw=gray!10},
      ]
      \addplot[blue,mark=o] table[x=wl, y=B]{\loadedtable};
	  \addplot[gray, mark=asterisk] table[x=wl, y=P]{\loadedtable};
	  \addplot[green, mark=otimes] table[x=wl, y=S]{\loadedtable};
	  \addplot[orange, mark=*] table[x=wl, y=SN]{\loadedtable};
	  \addplot[mark=none, black, dashed] coordinates {(0,86.37) (52,86.37)};
      \end{axis}
      \begin{axis}[
      xmin=0, xmax=55,scale = 0.7,
      ymax=1,
      hide x axis,
      axis y line=right,
      ylabel={KL div.},
      ylabel style={yshift=7.5pt,},
      table/col sep=comma,
      y axis style=red,
      ]
      \addplot[red, dashed, mark=otimes, mark size=1pt] table[x=x, y=y]{data/BCCD_KLdiv.csv};
      \end{axis}
    \end{tikzpicture}
    \end{minipage}
    \caption{Object detection results with and without noise to the point annotations. The right y-axis shows the KL divergence of $P_{100}$ from $P_{wl}$ where $P$ denotes the joint pdf of bounding boxes from a train set having $wl\%$ box-labelled images.}
    \label{fig:results}
\end{figure*}
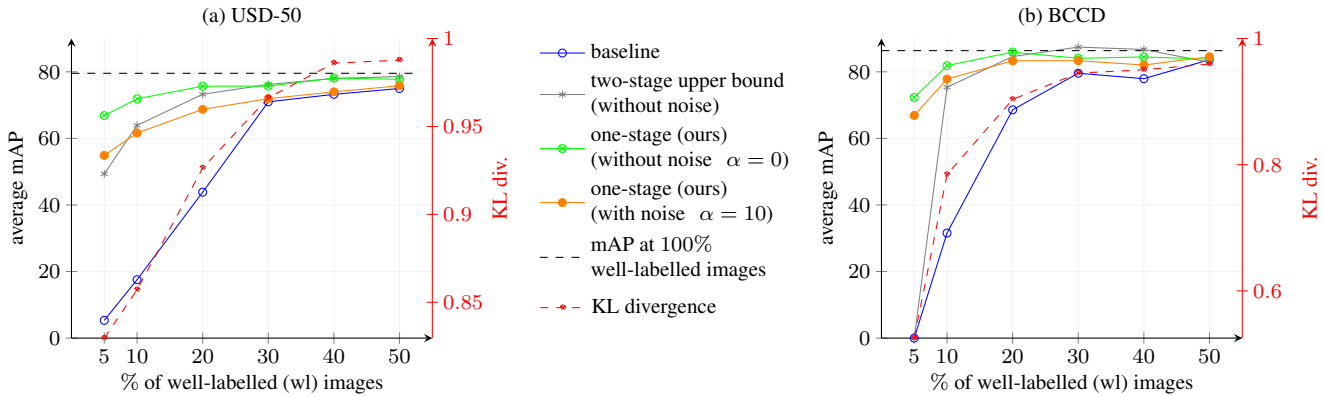

%% file: conclusion.tex
\section{Conclusion}

In this work, we propose a one-stage weakly semi-supervised method that jointly trains both box- and point-labelled object instances. Our method is simple to implement. The experiments on two datasets confirm that our one-stage method outperforms a class of two-stage methods.